\documentclass[sigconf]{acmart}
\setcopyright{none}

\usepackage{booktabs}
\usepackage{multirow}
\usepackage{graphicx}
\usepackage{enumitem}
\usepackage{tikz}
\usetikzlibrary{arrows.meta,positioning,shapes.geometric,calc,fit,backgrounds}

\renewcommand\footnotetextcopyrightpermission[1]{}
\acmConference[UDM@KDD '26]{The 5th Workshop on Uncertainty Reasoning and Quantification in Decision Making}{August 2026}{Jeju, Korea}
\acmYear{2026}
\copyrightyear{2026}

\author{%
  Vignesh Ram Nithin Kappagantula\textsuperscript{*},
  Shayan Hassantabar\textsuperscript{*},
  Samuel Simpson, and Golnaz Moallem
}
\email{{vkappagantula, shassantabar, a-sasimpson, gmoallem}@expediagroup.com}
\affiliation{%
  \institution{Expedia Group}
  \city{Seattle}
  \state{Washington}
  \country{USA}
}

\begin{document}


\title{Travel-Oriented Reasoning Large Language Model via Domain-Specific Knowledge Graphs}


\begin{abstract}

Large language models (LLMs) demonstrate broad reasoning abilities but struggle with accuracy and reliability in specialized domains such as travel, where reasoning depends on precise definitions, rules, and expert-defined conceptual frameworks, and where confident but unfounded outputs arise from a reasoning failure in which the model has not internalized the underlying domain graph rather than from missing domain knowledge alone. We propose a modular pipeline for building a travel-domain reasoning LLM grounded in an expert-designed knowledge graph (KG). Our pipeline integrates a travel KG that encodes domain entities and their relationships, a bottom-up construction procedure that walks the KG to produce multi-hop question answer (QA) pairs, a supervised fine-tuning stage that embeds the domain knowledge into a reasoning-capable LLM using the generated QA pairs as auditable reasoning traces, and a travel-domain benchmark dataset that measures the fine-tuned model's accuracy and calibration. We evaluate our approach using Qwen3-4B with LoRA adaptation. Our reasoning model achieves an $82.4\%$ exact match on the benchmark. This performance significantly outperforms the pretrained Qwen3-4B baseline at $22.4\%$. A calibration analysis decomposes the residual $17.57\%$ of errors into two distinct failure modes: an over-confident multi-label decoder that predicts both correct answers plus one spurious option on most dual-answer mistakes, and a smaller reasoning failure on single-answer questions where the supporting facts are present in the KG but the model fails to reconstruct the correct multi-hop path. This split confirms that explicit KG-grounded reasoning substantially improves the accuracy and uncertainty interpretation of LLMs in specialized domains, and isolates per-option calibration and trace-length-aware decoding as the next axes of improvement.

\end{abstract}



\maketitle
\renewcommand{\thefootnote}{*}%
\footnotetext{These authors contributed equally to this work.}%
\renewcommand{\thefootnote}{\arabic{footnote}}

\raggedbottom

\section{Introduction}
\label{sec:intro}

Large language models (LLMs) demonstrate remarkable reasoning capabilities across general domains ~\cite{wei2022chain, kojima2022large}. They successfully follow logical steps, generate complex explanations, and perform multi-hop reasoning. We generally drive this progress by pretraining models on vast text corpora and fine-tuning them on diverse instruction datasets. However, when we apply these models to highly specific fields such as the travel domain, they frequently struggle to maintain accuracy and reliability. This challenge arises because travel domain reasoning requires more than general logic. It demands strict adherence to the precise definitions, rules, and conceptual frameworks that govern the domain.

This discrepancy highlights a critical distinction between general reasoning and domain-specific reasoning. General reasoning relies on broad linguistic patterns, whereas domain reasoning depends on local ontologies, explicit rules, and contextual norms that dictate how we apply and interpret knowledge \cite{pan2024unifying}. When we fail to ground LLMs in these domain-specific structures, the models inevitably hallucinate. Therefore, to improve travel domain reasoning, we must ground LLMs in authoritative travel knowledge and structured logic, ensuring we generate consistent, accurate, and trustworthy outputs.

Currently, the standard methodology for teaching reasoning capabilities relies on a top-down approach. We typically expect models to learn general abstractions from massive collections of facts and statements using large-scale pretraining, reinforcement learning, and inference-time compute. Unfortunately, this top-down solution yields suboptimal learning efficiency in specialized areas like the travel domain. High-quality travel data is scarce, making grounded reasoning rooted in structured domain knowledge absolutely essential.

To overcome these limitations, we propose a bottom-up approach that starts with core travel knowledge and builds upward to develop advanced reasoning capabilities. We first encode the domain's glossary, rules, and data structures. We then connect the model to reliable sources and tools that enforce factual grounding. We implement a curriculum progression that guides the model from basic definitions to applied reasoning, teaching it to reason accurately step by step. Ultimately, we force the LLM to transition from merely producing likely-sounding answers to generating verifiable, rule-consistent reasoning. By actively grounding the models in foundational knowledge and structured logic, we successfully reduce hallucinations and transform them into reliable domain assistants.

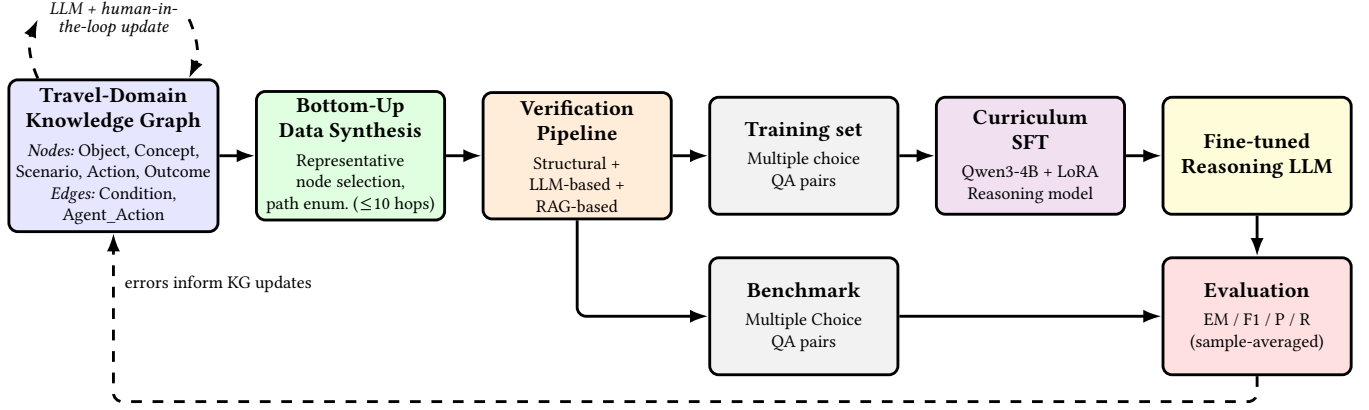
\begin{figure*}[t]
\centering
\resizebox{\textwidth}{!}{%
\begin{tikzpicture}[
  >={Latex[length=1.8mm,width=1.3mm]},
  every node/.style={font=\scriptsize, align=center},
  stage/.style={rectangle, rounded corners=2.5pt, draw, thick,
                minimum width=1.85cm, minimum height=1.15cm, inner sep=2pt},
  kg/.style    ={stage, fill=blue!10},
  synth/.style ={stage, fill=green!12},
  verif/.style ={stage, fill=orange!15},
  data/.style  ={stage, fill=gray!10},
  train/.style ={stage, fill=violet!12},
  model/.style ={stage, fill=yellow!18},
  eval/.style  ={stage, fill=red!12},
  arr/.style      ={->, thick, rounded corners=2.5pt},
  feedback/.style ={->, thick, dashed, rounded corners=2.5pt},
  node distance=3.5mm and 3.5mm
]

\node[font=\tiny, align=center] (upd)
  {\emph{LLM + human-in-}\\[-1pt]\emph{the-loop update}};

\node[kg, below=3mm of upd] (kg)
  {\textbf{Travel-Domain}\\\textbf{Knowledge Graph}\\[1pt]
   \tiny \emph{Nodes:} Object, Concept,\\[-1pt]
   \tiny Scenario, Action, Outcome\\[-1pt]
   \tiny \emph{Edges:} Condition,\\[-1pt]
   \tiny Agent\_Action};

\node[synth, right=of kg] (synth)
  {\textbf{Bottom-Up}\\\textbf{Data Synthesis}\\[1pt]
   \tiny Representative\\[-1pt]
   \tiny node selection,\\[-1pt]
   \tiny path enum.\ ($\leq$10 hops)};

\node[verif, right=of synth] (verif)
  {\textbf{Verification}\\\textbf{Pipeline}\\[1pt]
   \tiny Structural +\\[-1pt]
   \tiny LLM-based +\\[-1pt]
   \tiny RAG-based};

\node[data, right=of verif] (tr)
  {\textbf{Training set}\\[1pt]
   \tiny Multiple choice\\[-1pt]
   \tiny QA pairs};

\node[train, right=of tr] (sft)
  {\textbf{Curriculum}\\\textbf{SFT}\\[1pt]
   \tiny Qwen3-4B + LoRA\\[-1pt]
   \tiny Reasoning model};

\node[model, right=of sft] (fm)
  {\textbf{Fine-tuned}\\\textbf{Reasoning LLM}};

\node[data, below=4mm of tr] (bm)
  {\textbf{Benchmark}\\[1pt]
   \tiny Multiple Choice\\[-1pt]
   \tiny QA pairs};

\node[eval, below=4mm of fm] (ev)
  {\textbf{Evaluation}\\[1pt]
   \tiny EM / F1 / P / R\\[-1pt]
   \tiny (sample-averaged)};

\draw[arr] (kg)    -- (synth);
\draw[arr] (synth) -- (verif);
\draw[arr] (verif) -- (tr);
\draw[arr] (tr)    -- (sft);
\draw[arr] (sft)   -- (fm);

\draw[arr] (verif.south) |- (bm.west);

\draw[arr] (bm) -- (ev);
\draw[arr] (fm) -- (ev);

\draw[feedback] (kg.north -| upd.west) to[bend left=25] (upd.west);
\draw[feedback] (upd.east) to[bend left=25] (kg.north -| upd.east);

\draw[feedback]
    (ev.south) -- ++(0,-2.5mm) -| (kg.south);
\node[font=\tiny, anchor=west]
  at ($(kg.south)+(0mm,-5mm)$)
  {errors inform KG updates};
\end{tikzpicture}}
\caption{An overview of training a reasoning model using domain-specific knowledge graphs.}
\label{fig:framework}
\end{figure*}

In this paper, we introduce a novel, bottom-up reasoning framework comprising four deeply integrated modules, as shown in Fig.~\ref{fig:framework}:

\begin{itemize}[nosep,leftmargin=*]
\item \textbf{Travel Domain Knowledge Graph (KG)}: We encode foundational travel domain information into a robust, structured knowledge graph.
\item \textbf{Bottom-Up Travel Knowledge Construction and Verification} : We generate structured training data directly from the KG. To do this, we traverse the graph, synthesize instructions, formulate multiple-choice scenarios, and verify the factual accuracy of the data.
\item \textbf{Bottom-Up Curriculum Learning} : We train the travel domain reasoning model by systematically feeding it knowledge acquired from the KG, gradually increasing the complexity of the reasoning tasks.
\item \textbf{Travel Domain Benchmark} : We provide a comprehensive new benchmark that evaluates both the factual knowledge and the specific reasoning capabilities of LLMs operating within the travel sector.
\end{itemize}

The remainder of the paper proceeds as follows. Section~\ref{sec:related} reviews related work on reasoning in LLMs, KG grounded language models, and parameter efficient domain adaptation. Section~\ref{sec:kg} describes the travel domain ontology and the structure of the knowledge graph. Section~\ref{sec:construction} details our bottom-up procedure for synthesizing and verifying multi-answer multiple-choice questions from KG paths. Section~\ref{sec:training} presents the supervised fine-tuning setup of the Qwen3-4B model. Section~\ref{sec:benchmark} introduces the held-out benchmark, defines the evaluation metrics, reports our results, and characterizes the model's residual errors as a calibration problem on multi-answer questions. Finally Section~\ref{sec:conclusion} concludes the paper.

\section{Related Work}
\label{sec:related}

In this section, we review the prior work in three
related areas: reasoning in large language models, unification of KGs and LLMs, and domain specific models through instruction fine-tuning. 

\textbf{Reasoning in large language models.} Chain-of-thought (CoT) prompting elicits multi-step reasoning by asking models to verbalize intermediate steps~\cite{wei2022chain}, and models display such capabilities in a zero-shot setting when prompted appropriately~\cite{kojima2022large}. Self-consistency~\cite{wang2023selfconsistency} further improves reliability by sampling multiple reasoning paths and marginalizing over answers. More recently, reasoning-specialized models such as DeepSeek-R1~\cite{deepseekai2025r1} and the Qwen3 family~\cite{qwen3} show that distilled or RL-trained CoT can be internalized into smaller dense models. Our work complements this line by grounding CoT traces in an authoritative domain KG rather than relying solely on model-generated rationales, thereby reducing hallucinations in the specialized travel domain.

\textbf{Knowledge graphs and LLMs.} A growing body of work combines structured knowledge with neural language models. Pan et al.~\cite{pan2024unifying} provide a roadmap unifying KGs and LLMs along three axes: KG-enhanced LLMs, LLM-augmented KGs, and their synergistic use. Retrieval-augmented generation (RAG)~\cite{lewis2020rag} conditions generation on retrieved passages, while QA-GNN~\cite{yasunaga2021qagnn} and related methods reason jointly over text and a KG subgraph. Think-on-Graph~\cite{sun2024thinkongraph} performs explicit beam search over a KG at inference time to support multi-hop QA. These approaches typically augment inference with external structure; in contrast, we use the KG offline to synthesize verified training data and distill graph-grounded reasoning directly into model weights via SFT.

\textbf{Domain-specific and instruction fine-tuning.} Instruction tuning~\cite{ouyang2022instructgpt} and parameter-efficient adaptation have become standard for aligning LLMs to specialized tasks. LoRA~\cite{hu2022lora} introduces low-rank adapters that enable fine-tuning of large models with minimal compute, and QLoRA~\cite{dettmers2023qlora} extends this to quantized backbones. Domain specialization has yielded notable results in medicine (Med-PaLM~\cite{singhal2023medpalm}), mathematics (MAmmoTH~\cite{yue2024mammoth}), and code, typically by curating or synthesizing domain corpora. In the same way, our pipeline adopts a two-stage curriculum that moves from direct answers to reasoning-enhanced answers, what sets it apart is that each stage is generated from a maintained ontology, ensuring that every training example is structurally and logically valid by construction.

\section{Travel Domain Knowledge Graph}
\label{sec:kg}
In this section, we present the travel-domain knowledge graph that grounds every training example and reasoning trace later in the pipeline. We first describe the ontology that specifies how entities and their relations are represented, and we then describe how we populate that ontology from travel policy documents and the taxonomy that domain experts already use in practice.

\subsection{Ontology Design}

We focus on travel policy documents, in particular cancellation policies. 
We design a domain-specific ontology that captures the logic travel experts use when reasoning about these policies.
We represent every entity uniformly as a node, and we let hierarchy emerge from the relations between entities rather than from predefined categories. We then traverse and interpret the graph purely through those relations. 
Each node therefore means only what its edges say it means. The label \texttt{Property}, for instance, defaults to a real-estate reading in general English, but in our graph it carries only the meaning given by its travel-policy edges (the alternatives it offers, the amenities it houses, the refund conditions it satisfies), so that unrelated meaning never leaks in. For the same reason, concepts that sound alike, such as \texttt{Compensation} and \texttt{Property Refund}, remain distinct because they are identified by their connections rather than by their labels.

\subsection{Travel Domain KG Construction}

We derive the nodes and relationships from a careful reading of travel policy documents and the taxonomy used by domain experts.
Our goal is to enumerate every relevant object and record how it connects to the others.

\textbf{Node types} capture the general category of an object and encode the role it plays in an interaction. Each type fixes a specific kind of fact and constrains how we interpret the object in context. For example, we classify ``Email to vendor -- Supplier waiver request -- Customer has hotel approval -- Pre-travel'' as an \texttt{Action} node because it names an event in a customer-service interaction that can trigger subsequent steps, and we classify ``Refund under the property, supplier waiving'' as an \texttt{Outcome} node because it names a terminal state. By distinguishing these types, the ontology tells us what plausibly precedes or follows a given node and gives the graph a coherent structural and temporal scaffolding.

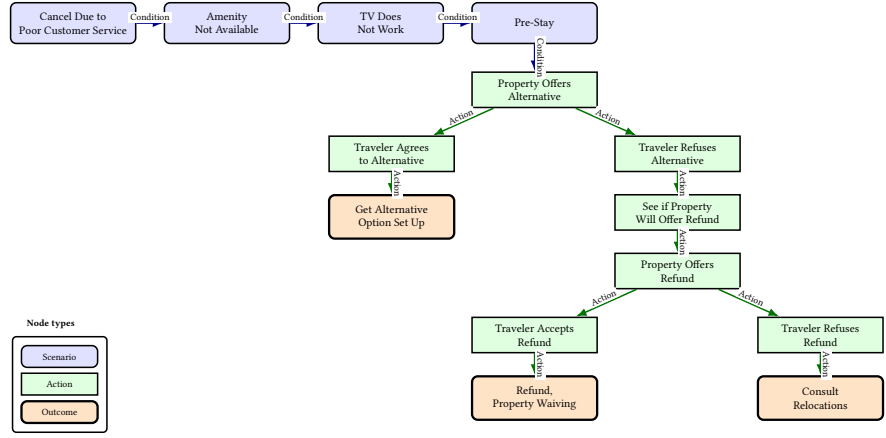
\begin{figure*}[t]
\centering
\resizebox{0.65\textwidth}{!}{%
\begin{tikzpicture}[
    >={Latex[length=2mm,width=1.5mm]},
    every node/.style={font=\scriptsize, align=center},
    scenario/.style={rectangle, rounded corners=3pt, draw, thick,
                     fill=blue!12, minimum width=2.3cm, minimum height=0.75cm, inner sep=1.5pt},
    condition/.style={rectangle, draw, thick,
                      fill=gray!10, minimum width=2.1cm, minimum height=0.65cm, inner sep=1.5pt},
    action/.style={rectangle, draw, thick,
                   fill=green!12, minimum width=2.3cm, minimum height=0.65cm, inner sep=1.5pt},
    outcome/.style={rectangle, rounded corners=3pt, draw, very thick,
                    fill=orange!22, minimum width=2.3cm, minimum height=0.8cm, inner sep=1.5pt},
    cond-edge/.style={->, thick, draw=blue!55!black},
    act-edge/.style ={->, thick, draw=green!45!black},
    elabel/.style={font=\tiny, sloped, anchor=south, midway,
                   inner sep=1pt, fill=white},
    node distance=4mm and 5mm
]

\node[scenario] (scn)            {Cancel Due to\\ Poor Customer Service};
\node[scenario, right=of scn] (c1) {Amenity\\ Not Available};
\node[scenario, right=of c1]  (c2) {TV Does\\ Not Work};
\node[scenario, right=of c2]  (c3) {Pre-Stay};

\node[action, below=5mm of c3] (a1) {Property Offers\\ Alternative};

\node[action, below left=5mm and 3mm of a1]  (a2a) {Traveler Agrees\\ to Alternative};
\node[action, below right=5mm and 3mm of a1] (a2b) {Traveler Refuses\\ Alternative};

\node[outcome, below=of a2a] (out1) {Get Alternative\\ Option Set Up};
\node[action,  below=of a2b] (a3)   {See if Property\\ Will Offer Refund};

\node[action, below=of a3] (a4) {Property Offers\\ Refund};

\node[action, below left=5mm and 3mm of a4]  (a5a) {Traveler Accepts\\ Refund};
\node[action, below right=5mm and 3mm of a4] (a5b) {Traveler Refuses\\ Refund};

\node[outcome, below=of a5a] (out2) {Refund,\\ Property Waiving};
\node[outcome, below=of a5b] (out3) {Consult\\ Relocations};


\draw[cond-edge] (scn) -- (c1) node[elabel] {Condition};
\draw[cond-edge] (c1)  -- (c2) node[elabel] {Condition};
\draw[cond-edge] (c2)  -- (c3) node[elabel] {Condition};
\draw[cond-edge] (c3)  -- (a1) node[elabel] {Condition};

\draw[act-edge]  (a1)  -- (a2a) node[elabel] {Action};
\draw[act-edge]  (a1)  -- (a2b) node[elabel] {Action};
\draw[act-edge]  (a2a) -- (out1) node[elabel] {Action};
\draw[act-edge]  (a2b) -- (a3)   node[elabel] {Action};
\draw[act-edge]  (a3)  -- (a4)   node[elabel] {Action};
\draw[act-edge]  (a4)  -- (a5a)  node[elabel] {Action};
\draw[act-edge]  (a4)  -- (a5b)  node[elabel] {Action};
\draw[act-edge]  (a5a) -- (out2) node[elabel] {Action};
\draw[act-edge]  (a5b) -- (out3) node[elabel] {Action};

\node[scenario,  minimum width=1.4cm, minimum height=4mm, font=\tiny,
      anchor=north west]
      at ([xshift=2mm, yshift=14mm]current bounding box.south west) (lg-scn)  {Scenario};
\node[action,    minimum width=1.4cm, minimum height=4mm, font=\tiny,
      below=0.8mm of lg-scn]  (lg-act)  {Action};
\node[outcome,   minimum width=1.4cm, minimum height=4mm, font=\tiny,
      below=0.8mm of lg-act]  (lg-out)  {Outcome};

\draw[thick, rounded corners=2pt]
  ([xshift=-1.5mm, yshift=1.5mm]lg-scn.north west)
  rectangle
  ([xshift=1.5mm, yshift=-1.5mm]lg-out.south east);

\node[font=\tiny\bfseries, anchor=north west]
  at ([xshift=2mm, yshift=22mm]current bounding box.south west) (lg-title) {Node types};

\end{tikzpicture}}
\caption{Local subgraph of the travel-domain KG anchored at \emph{Cancel Due to Poor Customer Service}, with node fill encoding ontology type and edge color separating \texttt{Condition} (blue) from \texttt{Action} (green) edges.}
\label{fig:kg-example}
\end{figure*}

\textbf{Edge types} encode the specific relationships between nodes. \texttt{Condition} edges specialize a scenario into a more specific scenario, and \texttt{Action} edges advance the interaction through an explicit step taken by the agent. Fig.~\ref{fig:kg-example} shows a representative subgraph anchored at ``Cancel Due to Poor Customer Service''. A chain of \texttt{Condition} edges narrows the context to a pre-stay TV-amenity issue, and an \texttt{Action} edge then moves the interaction to ``Property Offers Alternative''. From there the traveler either accepts and the path terminates at ``Get Alternative Option Set Up'', or refuses and triggers a compensation chain that eventually reaches ``Refund, Property Waiving'' or ``Consult Relocations''. Once the traveler reports that the TV does not work, the earlier \texttt{Condition} nodes drop out without changing the rest of the graph, so the graph stays agnostic to where reasoning starts and consistently drives every event toward an outcome.

We build the initial KG manually from the ontology and internal travel documents, but the underlying domain knowledge changes over time. We therefore keep the graph current with a human-in-the-loop pipeline in which an LLM proposes candidate triples from new internal documents and a domain expert accepts, edits, or rejects each proposal before we merge it back into the KG.

\section{Bottom-Up Knowledge Construction and Verification}
\label{sec:construction}

We adopt a bottom-up approach to knowledge construction and verification in the travel domain by synthesizing structured and contextual understanding from granular knowledge graph elements. Instead of relying on predefined taxonomies, we learn by traversing node relationships, performing contextual reasoning, and generating instructions that produce interpretable and domain grounded knowledge. Fig.~\ref{fig:kg-example} shows a local subgraph anchored at one such scenario.

\textbf{Representative Node Selection}. We begin by selecting representative starting nodes through balanced sampling across node types. We exclude outcome nodes as starting points except for a small diagnostic subset. We partition the graph into outcome and non-outcome nodes, where outcome nodes represent terminal decisions and are identified either by explicit labeling or by the absence of outgoing edges. We retain only those non-outcome nodes that can reach at least one outcome node through a directed path, verified via reachability checks. This filtering ensures that every sampled path corresponds to a valid decision trajectory.

\textbf{Path Length Sampling}. We then sample path lengths to control the depth of reasoning. For each valid start and outcome pair, we generate simple directed paths up to a maximum depth of ten hops. We prioritize shorter paths and retain up to five paths per pair to balance diversity and efficiency. The resulting paths capture both local and multi-hop relationships.

\textbf{Termination Criteria}. We terminate traversal when a path reaches an outcome node. We enforce strict validity by retaining only paths that end at verified outcome nodes and by validating each step against the graph to ensure that every transition corresponds to a valid directed edge.

\textbf{Instruction Generation}. We convert each path into a natural language instruction by combining a query derived from the start node with varying levels of contextual information from intermediate nodes. We generate multiple variants using diverse templates such as direct questions, scenario based prompts, and conditional formulations. We control context at four levels ranging from no context to full context. Each instruction includes a reasoning trace that narrates the traversal and a final answer derived from the terminal outcome node.

\textbf{Multiple Choice Formulation}. We transform these instructions into multiple choice questions. The correct answer corresponds to the outcome node, while distractors are selected from other outcome nodes using type aware and semantically informed criteria to ensure plausibility. When multiple paths yield the same question with different valid outcomes, we consolidate them into multi-answer questions. We maintain an approximately balanced training set with similar proportions of questions that have one, two, or three correct answers.

\textbf{Verification and Quality Assurance}. We verify each (question, reasoning, answer) triplet for logical consistency and domain grounding. We apply both model based and retrieval augmented verification. The model checks that reasoning steps follow from the graph structure and that all structural constraints are satisfied, including valid answer formatting and consistency. Graph level validation ensures that every reasoning trace exactly matches a valid path. Retrieval augmented verification cross checks each instance against relevant domain documents to confirm correctness. This pipeline ensures that the final dataset shown in Table \ref{tab:datastats} is accurate, interpretable, and free of structural or logical errors.

\begin{table}[t]
\caption{Knowledge graph and synthesized dataset statistics.}
\label{tab:datastats}
\centering
\small
\begin{tabular}{@{}lr@{}}
\toprule
\textbf{Statistic} & \textbf{Count} \\
\midrule
\textbf{Training QA pairs} & \textbf{2{,}764} \\
\quad 1-correct (\texttt{single\_answer}) & \textbf{882} \\
\quad 2-correct (\texttt{dual\_answer}) & \textbf{882} \\
\quad 3-correct (\texttt{multi\_answer}) & \textbf{1{,}000} \\
\textbf{Benchmark QA pairs (\S\ref{sec:benchmark})} & \textbf{888} \\
\quad 1-correct & \textbf{718} \\
\quad 2-correct & \textbf{163} \\
\quad 3-correct & \textbf{7} \\
\bottomrule
\end{tabular}
\end{table}

\section{Bottom-Up Curriculum Learning}
\label{sec:training}

We fine-tune two LoRA adapted variants of Qwen3-4B, a four-billion-parameter causal language model, with adapters on both the attention and MLP projection modules. Both variants start from the same pretrained base and train independently under matched hyperparameters, so the only differences between them are the training data and Qwen3's thinking-mode flag. The direct-answer variant disables thinking mode and trains on question and answer pairs alone. The reasoning-enhanced variant enables thinking mode and trains on the same questions augmented with KG-derived reasoning traces.

\subsection{Direct-Answer Training}

For the direct-answer variant, we apply Qwen3's chat template with thinking mode disabled. The target assistant turn is a single line of the form FINAL ANSWER: <letters>, where <letters> is a comma-separated alphabetical list of one to three correct choice letters. The training objective is next-token cross-entropy over that line. This format forces the model to internalize domain reasoning implicitly, since it sees only the question, the choices, and the set of correct letters. At inference the model emits only the answer letters, so the reasoning stays entirely inside the weights.

\subsection{Reasoning-Enhanced Training}

For the reasoning variant, we apply Qwen3's chat template with thinking mode enabled. The target assistant turn carries a <think> block with the KG-derived reasoning trace, followed by the explicit path and the final answer line, as shown below. The training objective is next-token cross-entropy over the full span from the opening <think> tag to the final answer letter. We draw every target trace from the exact KG path that generated the training example, so the model's reasoning stays structurally faithful to the source graph.

\begin{verbatim} 
<think> [Reasoning trace] </think>
Path 1: start -> intermediate -> ... -> outcome 
FINAL ANSWER: a, c 
\end{verbatim}

\subsection{Training Dataset Properties}

A defining characteristic of this dataset is that the same question text can appear with different choice sets and therefore different correct answer positions, by design, since the KG produces multiple valid outcome paths from the same starting scenario. This means the model cannot memorize question-to-letter mappings; it must learn to read the specific choice texts and match them against domain knowledge, a much harder and more generalizable task. With one to three correct answers per question and three to four carefully constructed distractors, the model must learn precise discrimination between correct policy outcomes and semantically plausible alternatives.

We algorithmically draw hard negatives from nearby nodes in the knowledge graph. This ensures that distractors are semantically plausible within the travel domain. They represent real policy outcomes that could apply under different conditions but not match the specific scenario. Notably, many choice texts in the dataset appear as both a correct answer and a distractor across different questions, making rote memorization of ``correct'' or ``incorrect'' answer texts impossible.

\section{Travel Domain Benchmark}
\label{sec:benchmark}
To evaluate the trained travel-domain reasoning LLM, we developed a domain-specific benchmark designed to assess both factual knowledge and policy-mapping capabilities. The benchmark uses the held-out validation split ($888$ samples) from the KG-generated dataset with the following properties:
\begin{itemize}[nosep,leftmargin=*]
    \item We draw all $888$ validation samples drawn from the knowledge graph.
    \item We balance the answer positions across all choice letters, preventing positional shortcut learning.
    \item Each benchmark questions draw its choices from a pool of 91 distinct Outcome-node strings, 74 of which appear as both correct answer and distractor across different questions, so accuracy depends on truly understanding both the question and its candidate answers rather than on memorizing the mapping between them.
\end{itemize}

\subsection{Evaluation Metrics and Results}

For each benchmark question $q$, let $Y_q \subseteq {a,b,c,d,e}$ be the set of gold letters and $\hat{Y}_q$ the model's predicted set. We compute four set-based metrics per question, namely exact match (EM), precision ($P$), recall ($R$), and F1 score ($F1$), and we arithmetically average each metric across the $N=888$ benchmark questions to obtain sample-averaged multi-label scores.

\begin{equation}
\label{eq:metrics}
\scalebox{0.9}{$
\text{EM}_q = \mathbf{1}[\hat{Y}_q = Y_q],\;
P_q = \tfrac{|Y_q \cap \hat{Y}_q|}{|\hat{Y}_q|},\;
R_q = \tfrac{|Y_q \cap \hat{Y}_q|}{|Y_q|},\;
F1_q = \tfrac{2 P_q R_q}{P_q + R_q}
$}.
\end{equation}

with $P_q = 0$ when $\hat{Y}_q = \emptyset$ and $F1_q = 0$ when $P_q + R_q = 0$. Reported numbers are the means of these per-question values over the benchmark. 

We compare three configurations: the pretrained Qwen3-4B baseline with no fine-tuning, a direct-answer fine-tuned model (LoRA SFT without reasoning traces, LR=5e-5), and a reasoning-enhanced fine-tuned model (LoRA SFT with reasoning traces, LR=5e-5). Table~\ref{tab:results} reports both overall results across the 888-sample benchmark and a breakdown by the number of correct answers per question.

Direct-answer fine-tuning improves overall exact match from 22.4\% to 66.0\%, and adding chain-of-thought reasoning boosts it further to 82.4\%, a +16.4pp gain from explicit reasoning alone. The impact is strongest on single-answer questions (93.3\% vs.\ 76.6\%), where the model reconstructs the KG path step by step before answering. On multi-answer questions, reasoning improves F1 from 0.560 to 0.770 on dual-answer and from 0.705 to 0.924 on triple-answer questions.

\begin{table}[t]
\caption{Benchmark test set results (888 samples). Best model per subset in \textbf{bold}.}
\label{tab:results}
\centering
\small
\begin{tabular}{@{}llcccc@{}}
\toprule
\textbf{Subset} & \textbf{Model} & \textbf{EM} & \textbf{F1} & \textbf{P} & \textbf{R} \\
\midrule
\multirow{3}{*}{Overall ($n{=}888$)}
 & Pretrained Qwen3-4B & 22.41 & 0.42 & 0.39 & 0.50 \\
 & Direct (LR=5e-5)    & 65.99 & 0.73 & 0.72 & 0.74 \\
 & Reasoning (LR=5e-5) & \textbf{82.43} & \textbf{0.90} & \textbf{0.89} & \textbf{0.92} \\
\midrule
\multirow{3}{*}{$|Y|{=}1$ ($n{=}718$)}
 & Pretrained Qwen3-4B & 24.79 & 0.41 & 0.37 & 0.50 \\
 & Direct (LR=5e-5)    & 76.60 & 0.77 & 0.77 & 0.77 \\
 & Reasoning (LR=5e-5) & \textbf{93.31} & \textbf{0.93} & \textbf{0.93} & \textbf{0.93} \\
\midrule
\multirow{3}{*}{$|Y|{=}2$ ($n{=}163$)}
 & Pretrained Qwen3-4B & 12.88 & 0.47 & 0.48 & 0.50 \\
 & Direct (LR=5e-5)    & 20.86 & 0.56 & 0.52 & 0.63 \\
 & Reasoning (LR=5e-5) & \textbf{34.97} & \textbf{0.77} & \textbf{0.71} & \textbf{0.86} \\
\midrule
\multirow{3}{*}{$|Y|{=}3$ ($n{=}7$)}
 & Pretrained Qwen3-4B &  0.00 & 0.57 & 0.75 & 0.52 \\
 & Direct (LR=5e-5)    & 28.57 & 0.70 & 0.76 & 0.67 \\
 & Reasoning (LR=5e-5) & \textbf{71.43} & \textbf{0.92} & \textbf{0.95} & \textbf{0.90} \\
\bottomrule
\end{tabular}
\end{table}

\subsection{Calibration on Multi-Answer Questions}

Table~\ref{tab:calibration} decomposes the residual errors of all three models. KG-grounded reasoning closes most of the knowledge gap rather than reshaping it, but the reasoning model's remaining $17.57\%$ of errors splits into two distinct failure modes that respond to different fixes.

The first is a residual reasoning failure on single-answer questions. The model is right 93.3\% of the time on this slice, and 47 of its 48 errors are disjoint, where the predicted letter has no overlap with the ground truth. These disjoint predictions therefore do not reflect missing knowledge in the graph but a learning gap in the weights, where the model has not internalized the specific multi-hop reasoning that connects the question to its correct answer. Calibration does not help here, because the model is not over-confident on a known answer. It is confidently following the wrong reasoning path.

The second is over-confidence on dual-answer questions. The model is right only 35.0\% of the time, and 66 of its 106 errors (62.3\%) are strict supersets in which it predicts both correct answers plus one spurious option, the canonical signature of an over-confident multi-label decoder addressable by per-option calibration such as temperature scaling or conformal thresholding. Correct predictions traverse 2.2 hops on average while errors traverse 10.3, so a length-aware stopping rule on the trace would catch the over-traversal that produces these spurious extras.

\begin{table}[t]
\caption{Set-shape decomposition (\%) of predictions on the 888-question multi-answer benchmark. SFT models trained at LR=5e-5. \emph{Empty} counts predictions from which no $\{a,\dots,e\}$ letter could be parsed.}
\label{tab:calibration}
\centering
\small
\setlength{\tabcolsep}{3.5pt}
\begin{tabular}{@{}lcccccc@{}}
\toprule
\textbf{Model} & \textbf{Exact} & \textbf{Super} & \textbf{Subset} & \textbf{Disjoint} & \textbf{Partial} & \textbf{Empty} \\
\midrule
Pretrained    & 22.41 & 23.87 & 2.48 & 44.93 & 5.52 & 0.79 \\
Direct SFT    & 65.99 &  4.50 & 0.45 & 22.64 & 6.42 & 0.00 \\
Reasoning SFT & 82.43 &  7.55 & 0.11 &  5.97 & 3.94 & 0.00 \\
\bottomrule
\end{tabular}
\end{table}

\section{Conclusion}
\label{sec:conclusion}

We built a travel-domain reasoning LLM by grounding Qwen3-4B in an expert-designed knowledge graph, synthesizing verified multi-hop QA pairs from KG traversals, and fine-tuning with reasoning traces drawn from the same paths. The reasoning-enhanced model achieves a 60.02 point EM gain over the pretrained baseline and a 16.44 point gain over a direct-answer variant trained on the same data, evidence that explicit KG-grounded reasoning, not fine-tuning alone, drives the improvement. The residual $17.57\%$ errors split into $7.5\%$ multi-answer over-confidence and $6.0\%$ single-answer reasoning failures. We therefore plan to extend the framework with a calibrated multi-label decoding head, a confidence-thresholded stopping rule on the reasoning trace, and stronger supervision on the underrepresented single-answer reasoning patterns.

\bibliographystyle{ACM-Reference-Format}

\begin{thebibliography}{99}

\bibitem{wei2022chain}
Jason Wei, Xuezhi Wang, Dale Schuurmans, Maarten Bosma, Brian Ichter, Fei Xia, Ed Chi, Quoc~V. Le, and Denny Zhou.
\newblock Chain-of-thought prompting elicits reasoning in large language models.
\newblock In \emph{Advances in Neural Information Processing Systems}, 2022.

\bibitem{kojima2022large}
Takeshi Kojima, Shixiang~Shane Gu, Machel Reid, Yutaka Matsuo, and Yusuke Iwasawa.
\newblock Large language models are zero-shot reasoners.
\newblock In \emph{Advances in Neural Information Processing Systems}, 2022.

\bibitem{wang2023selfconsistency}
Xuezhi Wang, Jason Wei, Dale Schuurmans, Quoc Le, Ed Chi, Sharan Narang, Aakanksha Chowdhery, and Denny Zhou.
\newblock Self-consistency improves chain-of-thought reasoning in language models.
\newblock In \emph{International Conference on Learning Representations}, 2023.

\bibitem{deepseekai2025r1}
DeepSeek-AI.
\newblock DeepSeek-R1: Incentivizing reasoning capability in LLMs via reinforcement learning.
\newblock \emph{arXiv preprint arXiv:2501.12948}, 2025.

\bibitem{qwen3}
Qwen Team.
\newblock Qwen3 technical report.
\newblock \emph{arXiv preprint arXiv:2505.09388}, 2025.

\bibitem{pan2024unifying}
Shirui Pan, Linhao Luo, Yufei Wang, Chen Chen, Jiapu Wang, and Xindong Wu.
\newblock Unifying large language models and knowledge graphs: A roadmap.
\newblock \emph{IEEE Transactions on Knowledge and Data Engineering}, 36(7):3580--3599, 2024.

\bibitem{lewis2020rag}
Patrick Lewis, Ethan Perez, Aleksandra Piktus, Fabio Petroni, Vladimir Karpukhin, Naman Goyal, Heinrich K\"uttler, Mike Lewis, Wen-tau Yih, Tim Rockt\"aschel, Sebastian Riedel, and Douwe Kiela.
\newblock Retrieval-augmented generation for knowledge-intensive NLP tasks.
\newblock In \emph{Advances in Neural Information Processing Systems}, 2020.

\bibitem{yasunaga2021qagnn}
Michihiro Yasunaga, Hongyu Ren, Antoine Bosselut, Percy Liang, and Jure Leskovec.
\newblock QA-GNN: Reasoning with language models and knowledge graphs for question answering.
\newblock In \emph{Proceedings of NAACL}, 2021.

\bibitem{sun2024thinkongraph}
Jiashuo Sun, Chengjin Xu, Lumingyuan Tang, Saizhuo Wang, Chen Lin, Yeyun Gong, Lionel~M. Ni, Heung-Yeung Shum, and Jian Guo.
\newblock Think-on-Graph: Deep and responsible reasoning of large language model on knowledge graph.
\newblock In \emph{International Conference on Learning Representations}, 2024.

\bibitem{ouyang2022instructgpt}
Long Ouyang, Jeff Wu, Xu~Jiang, Diogo Almeida, Carroll~L. Wainwright, Pamela Mishkin, Chong Zhang, Sandhini Agarwal, Katarina Slama, Alex Ray, John Schulman, Jacob Hilton, Fraser Kelton, Luke Miller, Maddie Simens, Amanda Askell, Peter Welinder, Paul Christiano, Jan Leike, and Ryan Lowe.
\newblock Training language models to follow instructions with human feedback.
\newblock In \emph{Advances in Neural Information Processing Systems}, 2022.

\bibitem{hu2022lora}
Edward~J. Hu, Yelong Shen, Phillip Wallis, Zeyuan Allen-Zhu, Yuanzhi Li, Shean Wang, Lu~Wang, and Weizhu Chen.
\newblock LoRA: Low-rank adaptation of large language models.
\newblock In \emph{International Conference on Learning Representations}, 2022.

\bibitem{dettmers2023qlora}
Tim Dettmers, Artidoro Pagnoni, Ari Holtzman, and Luke Zettlemoyer.
\newblock QLoRA: Efficient finetuning of quantized LLMs.
\newblock In \emph{Advances in Neural Information Processing Systems}, 2023.

\bibitem{singhal2023medpalm}
Karan Singhal, Shekoofeh Azizi, Tao Tu, S.~Sara Mahdavi, Jason Wei, Hyung~Won Chung, et~al.
\newblock Large language models encode clinical knowledge.
\newblock \emph{Nature}, 620:172--180, 2023.

\bibitem{yue2024mammoth}
Xiang Yue, Xingwei Qu, Ge~Zhang, Yao Fu, Wenhao Huang, Huan Sun, Yu~Su, and Wenhu Chen.
\newblock MAmmoTH: Building math generalist models through hybrid instruction tuning.
\newblock In \emph{International Conference on Learning Representations}, 2024.

\end{thebibliography}

\end{document}